\documentclass[conference]{IEEEtran}

\IEEEoverridecommandlockouts
\usepackage{cite}
\usepackage[numbers,sort&compress]{natbib}

\usepackage[utf8]{inputenc}
\usepackage{amsmath,amssymb}
\usepackage{graphicx}
\usepackage{bm}
\usepackage{enumerate}
\usepackage{comment}
\usepackage{xcolor}
\usepackage{url}
\usepackage{color,soul}
\usepackage{subcaption}
\usepackage{float}
\usepackage{tablefootnote}

\definecolor{light-gray}{gray}{0.8}

\usepackage{amsmath,amssymb,amsfonts}
\usepackage{algorithmic}
\usepackage{graphicx}
\usepackage{textcomp}
\usepackage{xcolor}
\usepackage{subcaption}

\begin{document}

\title{Deception Detection from Linguistic and Physiological Data Streams Using Bimodal Convolutional Neural Networks \\
}

\author{
\small 

\begin{tabular}[t]{c@{\extracolsep{8em}}c} 

1\textsuperscript{st} Panfeng Li\footnotemark & 2\textsuperscript{nd} Mohamed Abouelenien \\
\textit{Department of Electrical and Computer Engineering} & \textit{Department of Computer Science} \\
\textit{University of Michigan} & \textit{University of Michigan} \\
Ann Arbor, USA & Ann Arbor, USA \\
pfli@umich.edu & zmohamedra@umich.edu \\

\\

3\textsuperscript{rd} Rada Mihalcea & 4\textsuperscript{th} Zhicheng Ding \\
\textit{Department of Computer Science} & \textit{Fu Foundation School of Engineering and Applied Science} \\
\textit{University of Michigan} & \textit{Columbia University} \\
Ann Arbor, USA & New York, USA \\
mihalcea@umich.edu & zhicheng.ding@columbia.edu \\

\\
5\textsuperscript{th} Qikai Yang & 6\textsuperscript{th} Yiming Zhou \\
\textit{Department of Computer Science} & \textit{Department of Engineer Sciences} \\
\textit{University of Illinois Urbana-Champaign} & \textit{Saarland University of Applied Science} \\
Urbana, USA & Saarland, Germany \\
qikaiy2@illinois.edu & yiming.zhou@htwsaar.de \\

\end{tabular}

}

\maketitle

\begin{abstract}
  Deception detection is gaining increasing interest due to ethical and security concerns. This paper explores the application of convolutional neural networks for the purpose of multimodal deception detection. We use a dataset built by interviewing 104 subjects about two topics, with one truthful and one falsified response from each subject about each topic. In particular, we make three main contributions. First, we extract linguistic and physiological features from this data to train and construct the neural network models. Second, we propose a fused convolutional neural network model using both modalities in order to achieve an improved overall performance. Third, we compare our new approach with earlier methods designed for multimodal deception detection. We find that our system outperforms regular classification methods; our results indicate the feasibility of using neural networks for deception detection even in the presence of limited amounts of data.
\end{abstract}


\section{INTRODUCTION}
\label{sec:intro}  

Deception detection has been a topic of interest across many research fields -- ranging from psychology~\cite{DePaulo:03} to computer science~\cite{Ott:11}. With an ever-growing accessibility to multimodal media, for instance social media like YouTube and Snapchat, the detection of deceit based on multimodal data becomes increasingly necessary.

While deception detection is widely used in police interrogation, law enforcement, and employee security screening, the methods used often have a large time-requirement and rely highly upon physiological sensors and human experts, leading to bias and poor accuracy \cite{DePaulo:06}. There have been efforts to eliminate the need of human experts and introduce automated approaches. Machine learning methods have been used for the purpose of deception detection in the past, and efforts have been made to leverage multiple modalities to make predictions on the truthfulness of unseen data~\cite{Ott:11}.

These previous studies relied either on a single modality or on integrated multiple modalities in order to detect deceit using regular classification methods. The usage of a single modality might not provide enough information in order to detect deceit. On the other hand, the usage of multiple modalities means more information, and accordingly provides improved performance in many cases, reaching approximately 60-70\% accuracy~\cite{Abouelenien:14, Abouelenien:17}.

This implies that there is still room for improvement, and provides the opportunity to take advantage of the availability of multiple modalities to apply advanced learning techniques. Recent studies have shown that convolutional neural networks (CNNs)~\cite{yao2023improving} can improve the state-of-the-art performance on various tasks, including image analysis~\cite{Chen:20, zeng2024wordepth}, image classification~\cite{Krizhevsky:12, huiling_19}, localization and mapping~\cite{deng2023long, deng2024compact, Deng_2024_CVPR}, natural language processing~\cite{li2024contextualization, zhang2022can}, sentence classification~\cite{Kim:14, wang2023noiserobust}, which most recently inspired researchers' interests in utilizing deep learning into the deception detection problem. For instance,~\cite{Sun:16} implemented a fake review detection model using CNNs. However, a single modality was used to construct the network. An additional concern with the usage of multimodal data is the difficulty of collecting such data compared to a single modality. This fact causes the size of multimodal datasets to be limited, which may negatively affect the performance of deep learning methods, which do traditionally use very large datasets for training.

This paper addresses the problem of deception detection using multimodal neural networks. The paper makes three important contributions. First, we use neural networks to learn from two separate modalities, namely the linguistic and physiological modalities. Second, we construct a fused neural network that learns from both modalities, which to our knowledge has not been attempted before. Third, we compare our approach with earlier approaches that used regular machine learning techniques. Furthermore, we address the issues that arise using a CNN with a small training dataset by using a simple approach to solve the overfitting and large variance problems, namely using majority voting. We additionally devise a new procedure to deal with small datasets, including choosing an appropriate number of parameters as well as fixing the previous trained network weights to form a modality-wise training process. 

This paper is organized as follows. Section \ref{sec:related work} surveys some related work. Section \ref{sec:data} describes the dataset we used. Section \ref{sec:dl} illustrates the proposed deep learning approaches utilized for submodules as well as the whole framework. Section \ref{sec:exp} explains the experimental setup including data processing and feature extraction. Section \ref{sec:result} discusses our experimental results. Finally, concluding remarks and future work are provided in Section \ref{sec:conclusions}.

\section{Related Work}
\label{sec:related work}
Traditional methods mainly focused on the physiological indicators of deceit as the case with polygraph tests, such as blood pressure, respiration rate, and skin conductance. Different factors can affect the reliability of polygraphs including the fear of being perceived as a liar and the stress of being tested~\cite{NAP03}. Additionally countermeasures to fake innocence can be used, such as lying in the pretest questions and muscle tensing~\cite{Ganis11}. 

Another alternative to detect deception, for instance, is extracting features from the speaker's speech. Different studies have analyzed whether verbal cues were good indicators of deceptive behaviour. Examples of these clues included the speaker's pitch and speaking rate \cite{Hirschberg:05}. Other linguistic features have been extracted as well, such as the quantity, diversity, complexity, and specificity of messages, the word count and number of self-references, the keystroke dynamics and typing patterns, the corpus statistics and syntactic patterns, and the writing styles. There have also been efforts to use thermal imaging features for the purpose of detecting deception~\cite{Rajoub:14}.

The availability of multiple modalities offers the opportunity of extracting more information by considering the correspondences that exist naturally between multiple data sources \cite{Baltrusaitis:17}. In the domain of deception detection, feature fusion between linguistic, thermal, and physiological features has been explored for crowd-sourced data~\cite{Abouelenien:14}. 

For the purpose of automated detection of deceit, there has been research into applying traditional machine learning techniques. Such initiatives cast deception detection as a classification task, and use the available data to learn parameters for the model to be used for classification~\cite{Zhou:04b}.   

A more recent direction is the application of deep learning algorithms in this problem domain. Deep Learning methods have been used in natural language processing problems. For instance, Convolutional Neural Networks (CNNs) were used to produce state-of-the-art results on several problems in NLP~\cite{Kim:14}. Deep Learning for deception detection is more scant. Recent attempts were proposed to detect fake news and spam \cite{Wu:2017}. A new dataset for fake news has been benchmarked and released~\cite{Wang:17}. 

\section{Dataset}
\label{sec:data}
Our dataset includes two scenarios, namely ``Abortion" and ``Best Friend". The subjects were asked to sit comfortably on a chair in a lab and were connected to four physiological sensors including blood volume pulse, skin conductance, skin temperature, and abdominal respiration sensors. The participants were informed of the topic matter before each individual recording. In the two scenarios, subjects were allowed to speak freely first truthfully and then deceptively.

\textbf{Subjects.}
The multimodal dataset includes recordings collected from 104 students, including 53 females and 51 males. All subjects expressed themselves in English, had several ethnic backgrounds, and had an age range between approximately 20 and 35 years.

\textbf{Abortion.} In this scenario participants were asked to provide first a truthful and then a deceptive opinion about their feelings regarding abortion and whether they think it is right or wrong. The experimental session consisted of two independent recordings for each case.

\textbf{Best Friend.} In this scenario subjects were instructed to provide an honest description of their best friend, followed by a deceptive description about a person they cannot stand. In the deceptive response, they had to describe an individual they cannot stand as if he or she was their best friend. Hence, in both cases, the person was described positively.

\section{Bimodal CNNs}
\label{sec:dl}


Deep learning is an approach that has seen rapid growth in terms of popularity and usage, especially for classification tasks. We opted to use it for our classification task, where we aim at classifying the data as ``truthful'' or ``deceptive''. 

Our data is from two sources, namely the transcripts of the participants' responses, and the physiological data collected during the recordings. Accordingly, we utilize a linguistic CNN (LingCNN), a physiological CNN (PhysCNN), and a BiModal CNN network. The latter one fuses the previous two networks. In addition, a \textit{word2vec} model devised by~\cite{DBLP:journals/corr/abs-1301-3781} is used to transfer the transcripts to vectors as the input to our LingCNN.


Considering the size of our dataset is only $416$ instances, we set suitable hyperparameters, which correspond to reasonable numbers of weights in our networks. Furthermore, we utilize a modality-wise training fashion for our BiModal CNN, where we first train the linguistic and physiological CNNs, then use their output features as input for the BiModal CNN. We test our design using different experiments. 


\subsection{Vector Representations of Words}

Arbitrary discrete atomic encodings, as traditionally used in natural language processing tasks, provide little information about the semantic or syntactic relations between words that exist within the linguistic structure. Moreover, these representations lead to data sparsity, which leads to the need for large amounts of data in order to train statistical language models. Distributed vector representations of words have been shown to rectify a few of these problems, and have been shown to perform well on learning tasks for natural language processing. The distributed representations created by neural networks have some notion of linear translation. For our experiments, we use the \textit{word2vec} models \cite{DBLP:journals/corr/abs-1301-3781} to find the vector representations for our transcripts.

\noindent {\bf PhysCNN.} We construct a 1-Dimensional (1-D) CNN for the physiological modality. The inputs of the neural net consist of preprocessed physiological data with dimension of 32, the outputs are the classification results of the input samples. 

Firstly, the input data goes through the convolutional layer. We set three different filter sizes as $3,4,5$, which are the same with the ones in~\cite{Kim:14}. ReLU (Rectified Linear Unit) activation and max pooling are applied after convolution. All the pooled features are saved, concatenated, and flattened at the end.

We pass the flattened output through an added fully-connected layer, with a maximized activation, which provides our final prediction. Cross-entropy is used for training.

\noindent {\bf LingCNN.} We construct a convolutional model for our linguistic module, which is simplified from~\cite{Kim:14}'s TextCNN. In contrast to the PhysCNN model, this is a two-dimensional model. Similar to the cited paper, we chose filter sizes to be $3\times3$, $4\times4$, $5\times5$. 

\begin{figure}[ht]
    \centering
    \includegraphics[width=0.5\textwidth]{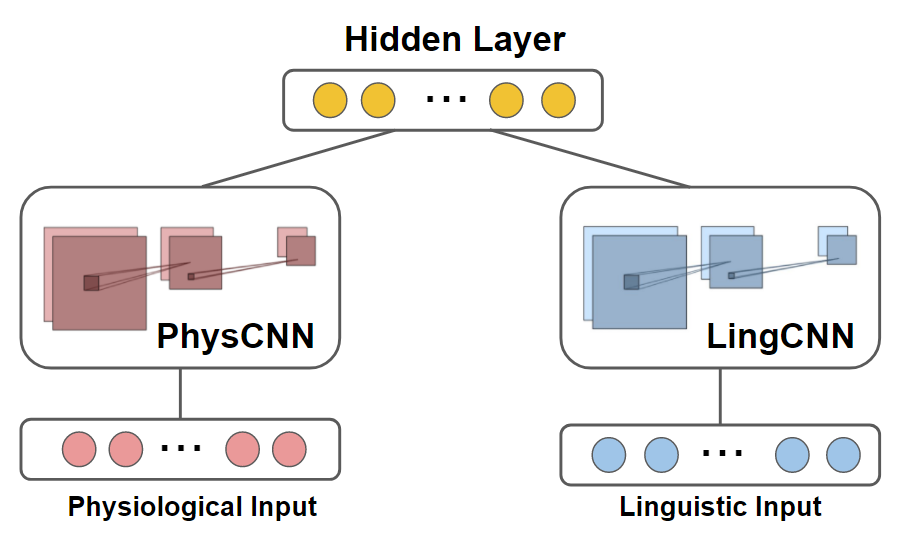}
    \caption{BiModal CNN} 
    \label{fig:bimodal}
\end{figure}

\noindent{\bf BiModal CNN.} The BiModal CNN represents a modality-wise fashion by first training the PhysCNN and LingCNN models. The relationship among them is shown in Figure~\ref{fig:bimodal}

\section{Experimental setup}
\label{sec:exp}

In this section we describe our experimental setup, including the data preprocessing as well as the training and testing procedures. 

\subsection{Data Preprocessing}
Here we describe the data preprocessing techniques on both of our modalities before passing them into our neural network models for feature extraction.

\subsubsection{Physiological Modality}

The physiological measurements are extracted at a rate of 2,048 samples per second using the Biograph Infinity Physiology suite. These features contain raw physiological measurements of the heart rate, skin conductance, respiration rate, and  skin temperature using four different sensors. Additionally, we compute their statistical descriptors including maximum and minimum values, means, power means, standard deviations, and mean amplitudes (epochs). The final physiological measurements set include a total of 59 physiological features that contain 40 features extracted from the raw measurement of the heart rate sensor, five skin conductance features, five skin temperature features, and seven respiration rate features. Furthermore, two measurements are extracted from the heart rate and the respiration rate sensors combined, namely, the mean and heart rate max-min difference, which represents a measure of breath to heart rate variability. 

We then simply average the values of the physiological data over the whole time period. The dimensions of the feature vectors are reduced from 59 to 32 following the application of Principal Component Analysis (PCA). PCA was used in order to reduce the features dimensions as well as the number of required weights in the network. Furthermore, our preliminary results indicated better performance following dimensionality reduction.

\subsubsection{Linguistic Modality}

Sentences in the transcripts were converted into word vectors in order to process the linguistic modality. To learn the representations of words, namely ``word embeddings'', we use the \textit{word2vec} model devised by~\cite{DBLP:journals/corr/abs-1301-3781}, where the training dataset is from Matt Mahoney. We set the embedding size, namely the length of word vectors as 32, similar to that of the physiological modality, and only keep the top 500 words with highest frequency in the text documents. Finally, we obtain a $500\times32$ word embedding matrix and a word dictionary, where each word corresponds to a unique value.

For each text transcript, we delete all non-verbal and non-numerical items and save the results as a transcript string, which is then transferred to a transcript vector through the dictionary described above. To unify the length of all the vectors for batch implementation in training and testing, we firstly identify the transcript vector(s), which have the maximum length $M$, and accordingly pad the remaining vectors with zeros. If a word does not exist in the dictionary, we replace it with a special notation as ``UNK'', which also corresponds to the value zero. Furthermore, each value in the transcript vectors is transferred to a word vector through a lookup operation on the previous embedding matrix. Hence, the transcripts are represented as arrays with dimension $M \times 32$.

\subsection{Training and Testing Procedures}

We randomly shuffle and split our dataset for training and testing with a ratio of $9:1$ and save the shuffled and split index. By using the same index, we are able to match the features from the two modalities, when integrated together.

For the linguistic and physiological modalities, the final predictions are obtained after applying a maximization function on the output scores of the network. The integrated network takes the output scores from linguistic and physiological modalities as input, and concatenates them as a single feature vector. The details of training and testing for the overall framework one-time are as follows: 

\begin{itemize}
    \item Train linguistic and physiological modalities once using all the training data.
    \item Fix the weights for the linguistic and physiological modalities and input the training and testing data to obtain the corresponding linguistic and physiological features for training and testing.
    \item Use the above training features as inputs for training the overall framework, and record the test results on testing features.
\end{itemize}

Specifically, we apply the majority voting method to determine the final predictions in order to address the overfitting and variance problem of the network. We record all the prediction results among a certain number of running times and decide the label for each test sample using the mode value. We also perform a stability analysis in Section \ref{sec:result}, which shows the majority voting method is effective and stable.

\section{Experimental Results and Discussion}
\label{sec:result}
Our entire dataset consists of 416 samples including the ``Abortion'' and ``Best Friend'' topics. We evaluate the performance of the features extracted from each of the two topics as well as both topics combined using the overall accuracy and class recall. Moreover, we compare the performance of individual modalities to that of their combination. Furthermore, we compare the performance of our proposed networks to that of learning using regular classifiers such as Decision Tree, Support Vector Machine (SVM), and Logistic Regression.

\subsection{Individual and integrated modalities}

\label{subsec:modalities}
\begin{figure}[ht]
    \centering
    \includegraphics[width=0.5\textwidth]{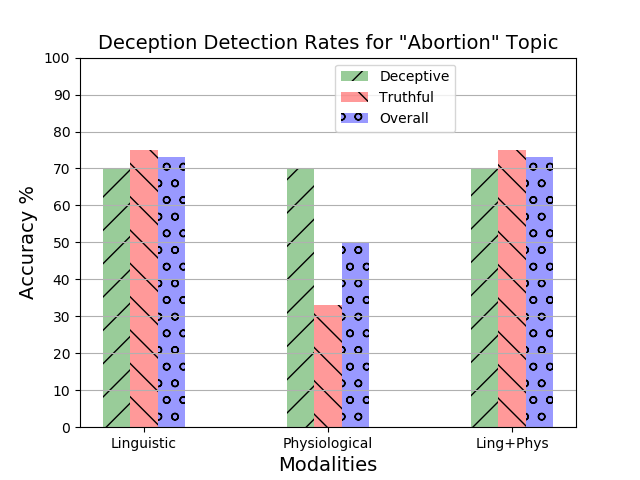}
    \caption{Deception recall, truthfulness recall, and overall accuracy percentages for individual and integrated modalities using features extracted from the ``Abortion'' topic} 
    \label{fig:abortion}
\end{figure}

Figure~\ref{fig:abortion} shows the deception and truthfulness recall in addition to the overall accuracy using different modalities for the ``Abortion'' topic. The figure indicates that overall the combination of linguistic and physiological modalities improves the performance as compared to the physiological modality. Specifically, while the physiological modality achieves the highest accuracy for the deception class, it attains the lowest truthful class accuracy, and is not performing as well as the linguistic modality considering the overall accuracy. The linguistic features exhibits close performance to the integrated modality. We may state that, for the ``Abortion'' topic, the combination of physiological modality with the linguistic one does not benefit our model. 

\begin{figure}[ht]
    \centering
    \includegraphics[width=0.5\textwidth]{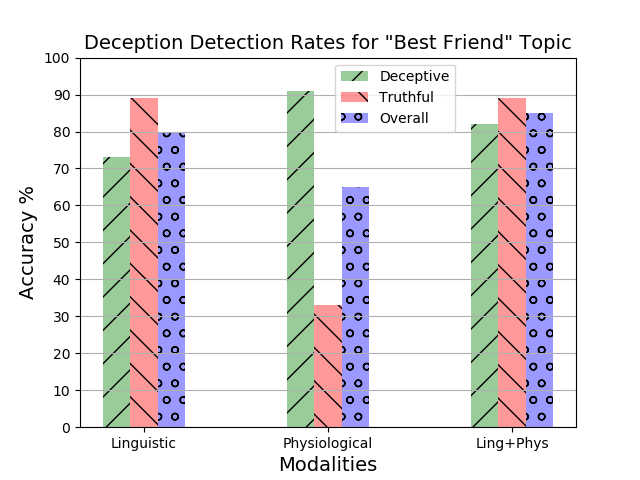}
    \caption{Deception recall, truthfulness recall, and overall accuracy percentages
 for individual and integrated modalities using features extracted from the ``Best Friend" topic} 
    \label{fig:best_friend}
\end{figure}

The performance of the features extracted from the ``Best Friend'' topic is significantly better than the first topic using different modalities as can be seen in Figure~\ref{fig:best_friend}. The overall accuracy using the linguistic modality reaches nearly $80 \%$ as compared to the approximately $70 \%$ achieved in the ``Abortion'' topic. Using the physiological modality, we compare $65 \%$ achieved for the ``Best Friend'' topic with $50 \%$ for ``Abortion'' topic on overall accuracy. The overall accuracy using both modalities indicates noticeable improvement compared to using individual modalities. 

\begin{figure}[ht]
    \centering
    \includegraphics[width=0.5\textwidth]{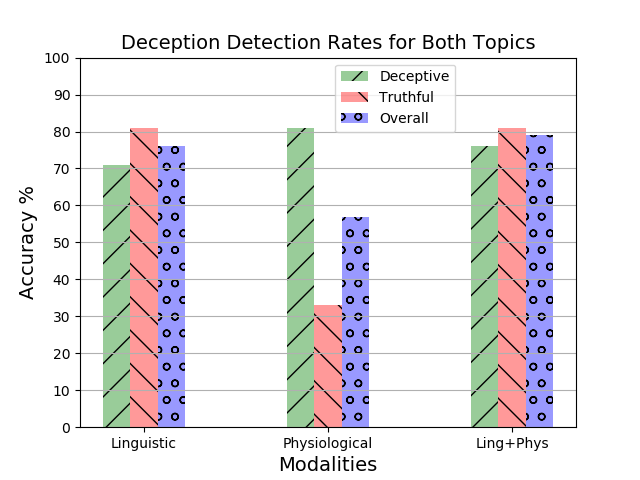}
    \caption{Deception recall, truthfulness recall, and overall accuracy percentages for individual and integrated modalities using features extracted from both the ``Abortion'' and ``Best Friend'' topic} 
    \label{fig:both_topics}
\end{figure}

Combining the two topics provides lower performance across all three modes of evaluation for all modalities. This may be rationalized by considering the fact that our model performed relatively poorly on the ``Abortion'' topic. As a result, the overall performance is slightly worse than that of ``Best Friend'' topic but better than ``Abortion'' topic. This can be seen in Figure ~\ref{fig:both_topics}.

In all three cases, we see that the detection rate of deceptive responses is better than that of the truthful one for the physiological modality. The reason behind this difference may be because the deceptive scenarios triggered more emotional arousal for the subjects, resulting in physiological patterns that were beneficial in training the networks. 
On the other hand, since the linguistic modality extracts semantic relations present in the same topic, the comparable performance for deceptive and truthful responses might be reasonable, as we train and test on data from the same topic.


\subsection{Cross-Topic Learning}     

\begin{figure}[ht]
    \centering
    \includegraphics[width=0.5\textwidth]{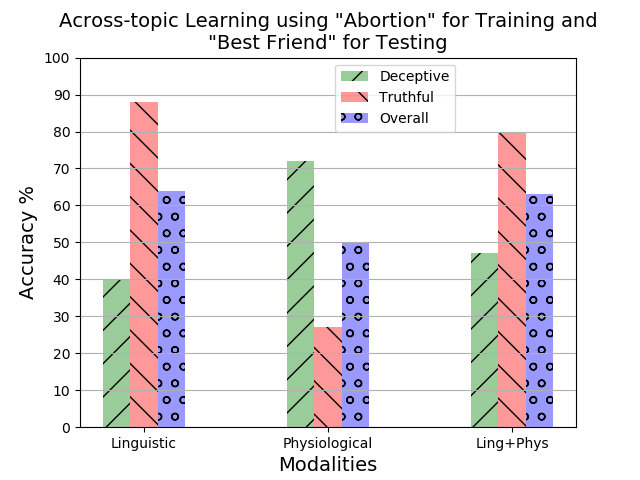}
    \caption{Deception, truthfulness, and overall accuracy percentages for individual and integrated modalities using across-topic learning. ``Abortion'' features are used for training and ``Best Friend'' features are used for testing} 
    \label{fig:cross_abortion}
\end{figure}

We analyze how well our model works on cross-topic deception detection. We train the model using the data from the ``Abortion'' topic and test on data from ``Best Friend'' topic. The results are presented in Figure~\ref{fig:cross_abortion}. 
The linguistic modality outperforms the physiological and the combined modalities on detecting truthful responses, but performs the worst of the three on deceptive responses. The overall accuracy of the integrated modality is similar with the one of linguistic and they both exceed $60 \%$.

This performance is flipped for the physiological modality, where we see the best performance is achieved using deceptive responses. Once again, this is likely because the physiological markers for deceptive responses are more indicative than those of the truthful responses. 

\begin{figure}[ht]
    \centering
    \includegraphics[width=0.5\textwidth]{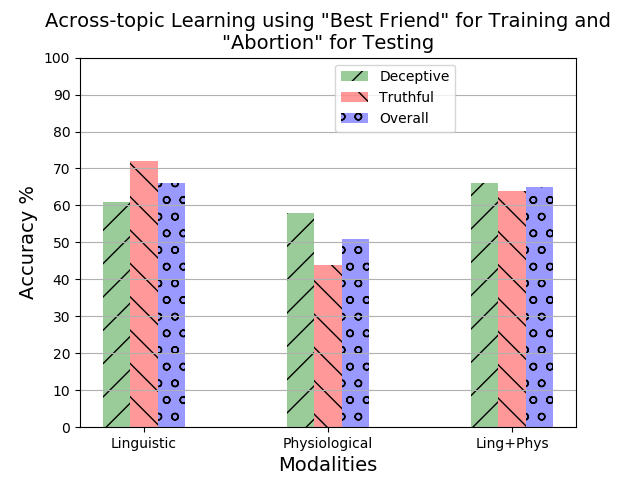}
    \caption{Deception recall, truthfulness recall, and overall accuracy percentages for individual and integrated modalities using across-topic learning. ``Best Friend'' features are used for training and ``Abortion'' features are used for testing} 
    \label{fig:cross_best_friend}
\end{figure}

In Figure~\ref{fig:cross_best_friend}, we can notice that while the trends are the same for linguistic and physiological modalities, the gaps between the recall figures for deceptive and truthful responses are significantly lower than the previous one. 
The results in this case indicate more stability regarding the truthful and deceptive classes performance compared to their performance in Figure~\ref{fig:cross_abortion}. This can be explained by having more domain-specific words in the ``Abortion'' topic, which affects the learning process. 

\begin{figure*}[ht]
    \centering
    \includegraphics[width=1.0\textwidth]{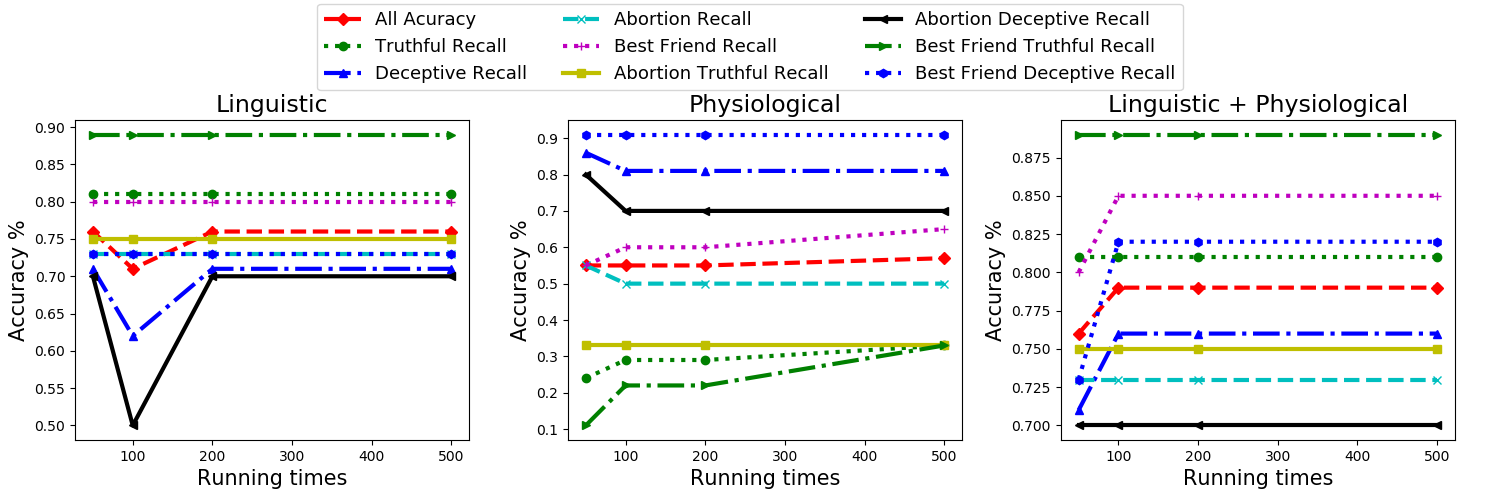}
    \caption{Accuracy results among different running times} 
    \label{fig:stability}
\end{figure*}

We may further compare these results with the ones discussed in subsection  \ref{subsec:modalities}. For the linguistic modality, the overall accuracy is lower for cross-topic learning, which indicates that the linguistic features are topic-dependent.

We also observe that the physiological modality, regardless of the topic used to train and testing consistently provides skewed results. Furthermore, training on ``Best Friend'' topic and testing on ``Abortion'' topic decreases the overall performance as compared to training and testing on the same ``Best Friend'' topic, but shows a very slight improvement as compared to training and testing on the same ``Abortion'' topic.

The combination of the two modules also does not perform as well regarding the overall accuracy for cross-topic learning as compared to the results in subsection \ref{subsec:modalities}.

\subsection{Stability Analysis}

Here we analyze the stability of our modalities. Since we determine our final predictions using majority voting among results from different running times, it is important to find the relationship between the accuracy and the number of running times. We tested the overall accuracy, deceptive recall, and truthful recall on individual topics and both topics combined. The results are shown in Figure~\ref{fig:stability}.

From Figure~\ref{fig:stability}, we can notice that the deceptive recall of the ``Abortion'' topic using the linguistic modality firstly decreases at 100 running times. The deceptive recall and overall accuracy also decrease accordingly. However, they quickly return to the normal level at 200 running times and stay consistent till 500 running times.

For the physiological modality, the truthful recall on the ``Best Friend'' topic and both topics combined is increasing when the running time goes from 50 to 500, while the deceptive recall on ``Best Friend'' and both topics is slightly decreasing. The ``All Accuracy" of the physiological modality stays consistent from 50 to 200 running times and increases slightly at 500 running times. 

For the integrated modalities, due to the increase of best friend deceptive and truthful recalls in the beginning, the recalls of the ``Best Friend'' topic and both topics increase. After 100 running times, all the accuracy figures remain unchanged, which indicates that the integrated modality is stable over running times despite the observed changes with the linguistic and physiological modalities. In conclusion, our models are stable after running times of 200.

\subsection{Compared with the Regular Models}

\begin{figure}[ht]
    \centering
    \includegraphics[width=0.5\textwidth]{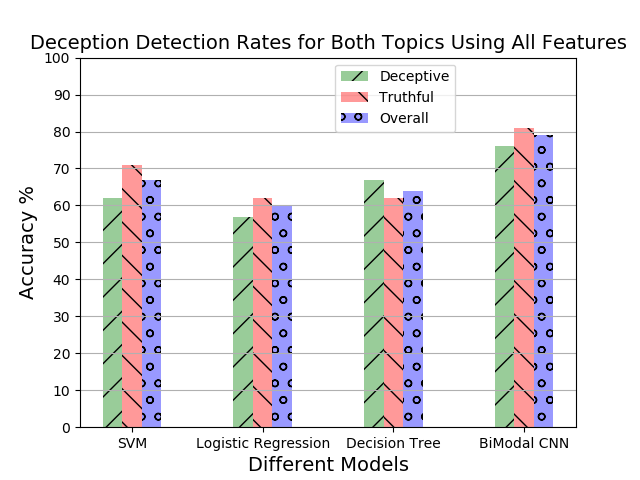}
    \caption{Comparison among different regular models and our BiModal CNN} 
    \label{fig:compare}
\end{figure}

We used the best multimodal systems for deception detection reported in a previous work ~\cite{Abouelenien:17} and compare their performance with ours. In those models, psycholinguistic lexicons and unigrams were used for linguistic features, while the paper used the same types of physiological features we utilized. In the end, the linguistic and physiological features were concatenated, and decision tree classifiers were used to give the final results. Here we also use SVM and logistic regression for classification. We compare results on the two topics combined -- ``Abortion'' and ``Best Friend'', and use both linguistic and physiological data. The results are shown in Figure~\ref{fig:compare}.

In our experiments, decision trees also used majority voting after a running them of 200. SVM and logistic regression did not need majority voting as their results remain stable over different running times. We note that, for all the different detection rates (deceptive, truthful and overall), our model performs better.

\section{Conclusions and Future Work}
\label{sec:conclusions}

This paper devised a method for using deep learning along with linguistic and physiological data for deception detection. The paper is the first, to our knowledge, to use multimodal data and deep learning to detect deception.

From the experimental results, we observed that the linguistic modality worked significantly better than the physiological modality. One of the reasons is that the linguistic modality used all the information in the transcripts, while the physiological modality simply averaged the data over the whole time period, which could result in loss of some physiological patterns in the learning process. 

It can also be noticed that in the majority of the cases, the bimodal network achieved better performance that the unimodal ones. This indicates that the proposed fused neural network can integrate and learn discriminative features from multimodal data, which results in improved and more reliable performance. 


For training and testing on the same topic, we note that by combining both modalities, the overall accuracy is higher than that obtained using the individual modalities. The same trend is observed for cross-topic learning, as well. We can therefore conclude that bimodal fusion has an overall advantageous effect over using individual modalities. This may be explained by considering that the fused network was provided by richer information using the two modalities.

Our experiments also indicated that cross-topic learning leads to a decrease in the performance for our model especially for the linguistic modality, which indicates that the performance is topic-dependent. 

For future work, we will consider performing a time-series analysis to potentially discover time-dependent relationships among the data. For the BiModal CNN, we will also extract different sizes of hidden layers from the LingCNN and PhysCNN, and then concatenate them to form new feature vectors.

\renewcommand{\bibfont}{\footnotesize}

\footnotesize{
\bibliographystyle{IEEEtran}
\bibliography{main}

\begin{thebibliography}{10}
\providecommand{\url}[1]{#1}
\csname url@samestyle\endcsname
\providecommand{\newblock}{\relax}
\providecommand{\bibinfo}[2]{#2}
\providecommand{\BIBentrySTDinterwordspacing}{\spaceskip=0pt\relax}
\providecommand{\BIBentryALTinterwordstretchfactor}{4}
\providecommand{\BIBentryALTinterwordspacing}{\spaceskip=\fontdimen2\font plus
\BIBentryALTinterwordstretchfactor\fontdimen3\font minus \fontdimen4\font\relax}
\providecommand{\BIBforeignlanguage}[2]{{%
\expandafter\ifx\csname l@#1\endcsname\relax
\typeout{** WARNING: IEEEtran.bst: No hyphenation pattern has been}%
\typeout{** loaded for the language `#1'. Using the pattern for}%
\typeout{** the default language instead.}%
\else
\language=\csname l@#1\endcsname
\fi
#2}}
\providecommand{\BIBdecl}{\relax}
\BIBdecl

\bibitem{DePaulo:03}
B.~M. DePaulo, J.~J. Lindsay, B.~E. Malone, L.~Muhlenbruck, K.~Charlton, and H.~Cooper, ``Cues to deception,'' \emph{Psychological Bulletin}, vol. 129, no.~1, pp. 74--118, 2003.

\bibitem{Ott:11}
M.~Ott, Y.~Choi, C.~Cardie, and J.~T. Hancock, ``Finding deceptive opinion spam by any stretch of the imagination,'' in \emph{Proceedings of the 49th Annual Meeting of the Association for Computational Linguistics: Human Language Technologies - Volume 1}, Stroudsburg, PA, USA, 2011, pp. 309--319.

\bibitem{DePaulo:06}
C.~F. Bond and B.~M. DePaulo, ``Accuracy of deception judgments,'' \emph{Personality and Social Psychology Review}, vol.~10, no.~3, pp. 214--234, 2006.

\bibitem{Abouelenien:14}
M.~Abouelenien, V.~P{\'e}rez-Rosas, R.~Mihalcea, and M.~Burzo, ``Deception detection using a multimodal approach,'' in \emph{Proceedings of the 16th International Conference on Multimodal Interaction}, 2014, pp. 58--65.

\bibitem{Abouelenien:17}
------, ``Detecting deceptive behavior via integration of discriminative features from multiple modalities,'' \emph{IEEE Transactions on Information Forensics and Security}, vol.~12, pp. 1042--1055, 2017.

\bibitem{yao2023improving}
J.~Yao, T.~Wu, and X.~Zhang, ``Improving depth gradientcontinuity in transformers: A comparative study on monocular depth estimation with cnn,'' \emph{arXiv preprint arXiv:2308.08333}, 2023.

\bibitem{Chen:20}
X.~Chen, A.~T.~Z. Kasgari, and W.~Saad, ``Deep learning for content-based personalized viewport prediction of 360-degree vr videos,'' \emph{IEEE Networking Letters}, vol.~2, no.~2, pp. 81--84, 2020.

\bibitem{zeng2024wordepth}
Z.~Zeng, D.~Wang, F.~Yang, H.~Park, Y.~Wu, S.~Soatto, B.-W. Hong, D.~Lao, and A.~Wong, ``Wordepth: Variational language prior for monocular depth estimation,'' \emph{arXiv preprint arXiv:2404.03635}, 2024.

\bibitem{Krizhevsky:12}
A.~Krizhevsky, I.~Sutskever, and G.~E. Hinton, ``Imagenet classification with deep convolutional neural networks,'' in \emph{Advances in Neural Information Processing Systems 25}, 2012, pp. 1097--1105.

\bibitem{huiling_19}
Y.~Tian, H.~Zhang, Y.~Jiang, P.~Li, and Y.~Li, ``A fusion feature for enhancing the performance of classification in working memory load with single-trial detection,'' \emph{IEEE Transactions on Neural Systems and Rehabilitation Engineering}, vol.~27, no.~10, pp. 1985--1993, 2019.

\bibitem{deng2023long}
T.~Deng, H.~Xie, J.~Wang, and W.~Chen, ``Long-term visual simultaneous localization and mapping: Using a bayesian persistence filter-based global map prediction,'' \emph{IEEE Robotics \& Automation Magazine}, vol.~30, no.~1, pp. 36--49, 2023.

\bibitem{deng2024compact}
T.~Deng, Y.~Chen, L.~Zhang, J.~Yang, S.~Yuan, D.~Wang, and W.~Chen, ``Compact 3d gaussian splatting for dense visual slam,'' \emph{arXiv preprint arXiv:2403.11247}, 2024.

\bibitem{Deng_2024_CVPR}
T.~Deng, G.~Shen, T.~Qin, J.~Wang, W.~Zhao, J.~Wang, D.~Wang, and W.~Chen, ``Plgslam: Progressive neural scene represenation with local to global bundle adjustment,'' in \emph{Proceedings of the IEEE/CVF Conference on Computer Vision and Pattern Recognition (CVPR)}, June 2024, pp. 19\,657--19\,666.

\bibitem{li2024contextualization}
D.~Li, Z.~Tan, T.~Chen, and H.~Liu, ``Contextualization distillation from large language model for knowledge graph completion,'' \emph{arXiv preprint arXiv:2402.01729}, 2024.

\bibitem{zhang2022can}
R.~Zhang, Z.~Zeng, Z.~Guo, and Y.~Li, ``Can language understand depth?'' in \emph{Proceedings of the 30th ACM International Conference on Multimedia}, 2022, pp. 6868--6874.

\bibitem{Kim:14}
Y.~Kim, ``Convolutional neural networks for sentence classification,'' in \emph{Proceedings of the 2014 Conference on Empirical Methods in Natural Language Processing, {EMNLP}}, 2014, pp. 1746--1751.

\bibitem{wang2023noiserobust}
D.~Li, Z.~Tan, T.~Chen, and H.~Liu, ``Noise-robust fine-tuning of pretrained language models via external guidance,'' \emph{arXiv preprint arXiv:2311.01108}, 2023.

\bibitem{Sun:16}
C.~Sun, Q.~Du, and G.~Tian, ``Exploiting product related review features for fake review detection,'' \emph{Mathematical Problems in Engineering}, 2016.

\bibitem{NAP03}
N.~R. Council, \emph{The Polygraph and Lie Detection}.\hskip 1em plus 0.5em minus 0.4em\relax Washington, DC: The National Academies Press, 2003.

\bibitem{Ganis11}
G.~Ganis, J.~P. Rosenfeld, J.~Meixner, R.~A. Kievit, and H.~E. Schendan, ``Lying in the scanner: Covert countermeasures disrupt deception detection by functional magnetic resonance imaging,'' \emph{NeuroImage}, vol.~55, no.~1, pp. 312--319, 2011.

\bibitem{Hirschberg:05}
J.~Hirschberg, S.~Benus, J.~M. Brenier, F.~Enos, S.~Friedman, S.~Gilman, C.~Girand, M.~Graciarena, A.~Kathol, L.~Michaelis \emph{et~al.}, ``Distinguishing deceptive from non-deceptive speech,'' in \emph{Ninth European Conference on Speech Communication and Technology}, 2005.

\bibitem{Rajoub:14}
B.~A. Rajoub and R.~Zwiggelaar, ``Thermal facial analysis for deception detection,'' \emph{Trans. Info. For. Sec.}, vol.~9, no.~6, pp. 1015--1023, Jun. 2014.

\bibitem{Baltrusaitis:17}
T.~Baltrusaitis, C.~Ahuja, and L.~Morency, ``Multimodal machine learning: {A} survey and taxonomy,'' \emph{CoRR}, vol. abs/1705.09406, 2017.

\bibitem{Zhou:04b}
L.~Zhou, J.~K. Burgoon, D.~P. Twitchell, T.~Qin, and J.~F. {Nunamaker, Jr.}, ``A comparison of classification methods for predicting deception in computer-mediated communication,'' \emph{Journal of Management Information Systems}, vol.~20, no.~4, pp. 139--166, 2004.

\bibitem{Wu:2017}
T.~Wu, S.~Liu, J.~Zhang, and Y.~Xiang, ``Twitter spam detection based on deep learning,'' in \emph{Proceedings of the Australasian Computer Science Week Multiconference}.\hskip 1em plus 0.5em minus 0.4em\relax New York, NY, USA: ACM, 2017, pp. 3:1--3:8.

\bibitem{Wang:17}
W.~Y. Wang, ``"liar, liar pants on fire": {A} new benchmark dataset for fake news detection,'' in \emph{Proceedings of the 55th Annual Meeting of the Association for Computational Linguistics, {ACL} 2017, Vancouver, Canada, July 30 - August 4, Volume 2: Short Papers}, 2017, pp. 422--426.

\bibitem{DBLP:journals/corr/abs-1301-3781}
T.~Mikolov, K.~Chen, G.~Corrado, and J.~Dean, ``Efficient estimation of word representations in vector space,'' \emph{arXiv preprint arXiv:1301.3781}, 2013.

\end{thebibliography}
}

\end{document}